\pdfoutput=1
\documentclass[11pt]{article}
\usepackage{acl}
\usepackage{marvosym}
\usepackage{times}
\usepackage{latexsym}
\usepackage[T1]{fontenc}
\usepackage[utf8]{inputenc}
\usepackage{microtype}
\usepackage{inconsolata}
\usepackage{graphicx}
\usepackage{amsmath}
\usepackage{amssymb}
\usepackage{multirow}
\usepackage{colortbl}
\usepackage{booktabs}
\usepackage{caption}
\usepackage{xcolor} 
\title{WindowKV: Task-Adaptive Group-Wise KV Cache Window Selection \\ for Efficient LLM Inference}

\author{
\\
Youhui Zuo\textsuperscript{1}, Sibo Wei\textsuperscript{2}, Chen Zhang\textsuperscript{1}, Zhuorui Liu\textsuperscript{1}, 
Wenpeng Lu\textsuperscript{2},
Dawei Song\textsuperscript{1 (\Letter)}\\
\textsuperscript{1}Beijing Institute of Technology, Beijing, China\\
 \textsuperscript{2}Qilu University of Technology (Shandong Academy of Sciences), Jinan, China\\
 \small{
 \textsuperscript{(\Letter)} Corresponding author.
 }
}

\begin{document}
\maketitle
\begin{abstract}
With the advancements in long-context inference capabilities of large language models (LLMs), the KV cache has become one of the foundational components. However, its substantial GPU memory consumption makes KV cache compression a key technique for enabling efficient LLM inference in industrial scenarios. While recent studies have focused on optimizing the memory occupied by the KV cache, they overlook two critical factors: preserving semantic coherence and considering task-specific characteristic during compression. To address these limitations, we propose a novel task-adaptive KV cache window selection method, \textbf{WindowKV}. WindowKV dynamically selects local semantic windows consisting of consecutive tokens, according to task-specific characteristics, ensuring the retained KV cache captures continuous, essential context. Additionally, we introduce an intra-group layer KV cache indices sharing strategy to reduce computational overhead, achieving a balance between performance and efficiency. We rigorously evaluate WindowKV on the LongBench benchmark, and the results demonstrate that it maintains a performance comparable to full KV cache retention while using only 12\% of the original KV cache, significantly reducing memory requirements. Furthermore, our method also achieves state-of-the-art results in the Needle-in-a-Haystack evaluation, highlighting its effectiveness and robustness.~\footnote{Our code is available at~\href{https://github.com/optim996/WindowKV}{GitHub}.}
\end{abstract}

\section{Introduction}
Tasks requiring long-context understanding, such as long-text reading comprehension~\cite{trivedi2022musique}, in-context learning~\cite{dong2024survey}, document summarization~\cite{huang2021efficient} and code completion~\cite{zheng2023codegeex}, have gained significant prominence in the era of LLMs. 
As a result, LLMs that are capable of processing extended context lengths have become increasingly prevalent~\cite{huang2023advancing}. For example, models like GPT-4 and DeepSeek-V3 support context lengths of up to 128K tokens, while Claude-3.5 and Yi extend this capability to 200K tokens~\cite{achiam2023gpt, liu2024deepseek_v3, fu2024data}. 
However, the self-attention mechanism in transformer architectures exhibits quadratic complexity with respect the context length~\cite{vaswani2017attention}, leading to significant increases in inference latency for long-context scenarios. One potential method to mitigate this latency is to cache the key and value (KV) states of previous tokens, thereby avoiding the recomputation of historical contexts~\cite{waddington2013kv}. Nevertheless, as both the input context length and the number of layers increase, the memory required to store the KV states increases substantially~\cite{luohekeep}. 
For instance, storing a KV cache for 100K tokens in the LLaMA2-7B model~\cite{touvron2023llama} demands over 50GB of memory, whereas a 2K token context requires less than 1GB~\cite{wu2024retrieval}. 
Overall, KV cache compression is essential to addressing issues such as memory demands, computational efficiency, energy consumption, and costs in LLMs, directly impacting their deployment and application effectiveness in industrial scenarios.

Recent studies have alleviated the aforementioned memory constraints by modifying attention architectures~\cite{shazeer2019fast, ainslie2023gqa, liu2024deepseek_v2}, or by implementing cross-layer sharing of the KV cache~\cite{brandon2024reducing, sun2024you, wu-tu-2024-layer}. However, these approaches require additional training. In contrast, another line of research has focused on compressing the KV cache during the inference phase. 
For example, some approaches evict the KV states of non-essential tokens under a fixed layer budget~\cite{zhang2023h2o, xiaoefficient, adnan2024keyformer, ge2024model}. However, these methods overlook differences in attention sparsity between layers.
PyramidInfer~\cite{yang2024pyramidinfer} and PyramidKV~\cite{cai2024pyramidkv} observe that dense attention is particularly prevalent in the lower layers, while sparse attention dominates in the higher layers~\cite{wang2023label}. They propose allocating varying KV cache sizes across layers to maintain a pyramid-like structure.

However, the aforementioned methods individually select KV cache tokens, and the selected discrete tokens disrupt the semantic consistency of the context. This also contradicts human reading habits. When processing long texts, humans typically do not read token by token but rather process information in windows~\cite{rayner1998eye}.
Moreover, these methods employ a uniform compression strategy across all tasks, limiting their task-specific adaptability and overall performance. In fact, based on human experience, the information processing approaches for different tasks vary significantly. For instance, in a question-answering task, this can be seen as an information localization task~\footnote{Information Localization Task: Identify the critical paragraphs within the given context, and then answer the relevant questions based on the information provided in these critical paragraphs.}, where the entire window is processed to capture comprehensive semantic details, thus facilitating accurate answer generation. In contrast, in a summarization task, the goal is information aggregation~\footnote{Information Aggregation Task: Extract essential information from each paragraph and compile it into a cohesive summary.}, where extract the most salient information from each window to generate a concise summary. The challenges outlined above motivate us to propose a task-adaptive KV cache window selection method. 
Additionally, to enhance computational efficiency, we integrated intra-group layer KV cache indices sharing strategy to better sustain the model's performance under constrained memory budgets.

In summary, our contributions are as follows:
\begin{itemize}
\item Different from previous KV cache compression methods that select discrete tokens, we introduce a task-adaptive window selection method, \textbf{WindowKV}. WindowKV dynamically compresses the KV cache based on task-specific characteristics while preserving the semantic coherence within local windows.
\item Additionally, we propose an intra-group layer KV cache indices sharing strategy to reduce computational overhead, achieving a balance between performance and efficiency.
\item Extensive experiments are conducted on the LongBench and Needle-in-a-Haystack benchmarks. The results demonstrate that WindowKV achieves the highest number of state-of-the-art results across various backbone LLMs and KV cache configurations on LongBench, while also surpassing other baseline methods on the Needle-in-a-Haystack test.
\end{itemize}

\begin{figure*}[t]
    \centering
    \includegraphics[width=1\textwidth]{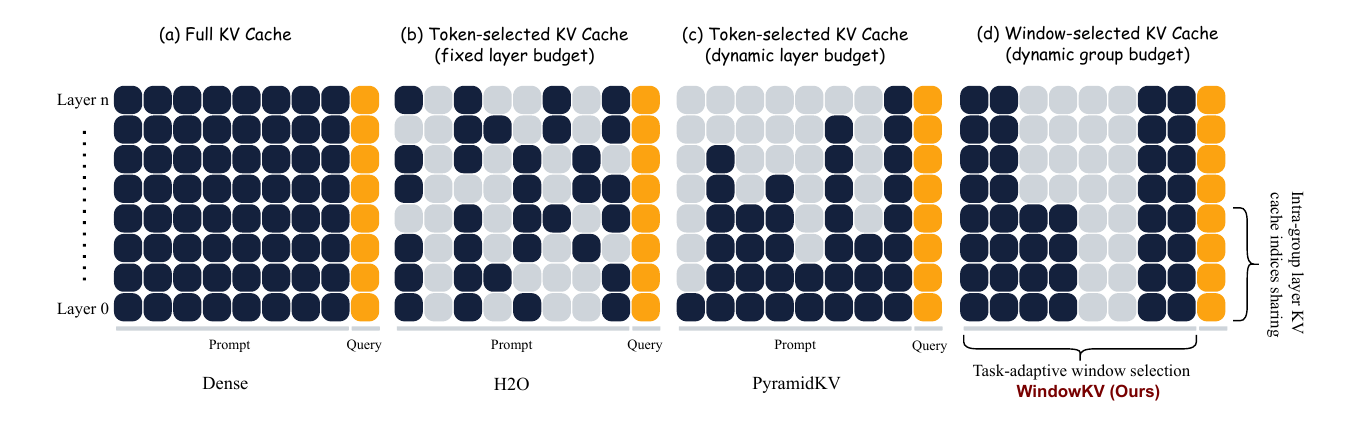}
    \caption{Comparison of WindowKV with state-of-the-art KV cache compression methods. (a) Full KV retains all tokens in the KV cache for each layer, with cache size growing linearly with input length. (b) H2O maintains a fixed cache size across layers, selecting tokens based on attention scores. (c) PyramidKV adopts a pyramid-shaped cache structure, allocating varying cache budgets to different layers. These methods uniformly apply token level selection strategies across all tasks. (d) WindowKV, in contrast, introduces a task-adaptive window selection method combined with intra-group layer KV cache indices sharing strategy, dynamically allocating group budgets across different groups.}
    \label{fig:model}
\end{figure*}

\section{Methodology} 
\subsection{Overview of KV Cache Compression}
In autoregressive transformer-based LLMs, the generation of the $i$-th token requires the attention module to access the KV states of all preceding $i-1$ tokens. To enhance the computational efficiency and avoid redundant calculations, these KV states are stored in the KV cache upon their initial computation. This caching mechanism significantly accelerates the inference process by eliminating the need for repeated computations. However, the KV cache can impose substantial memory demands, particularly for lengthy contexts. As a potential method, KV cache compression has been proposed to save memory while minimizing information loss as much as possible.

\subsection{WindowKV}
In this section, we introduce WindowKV, a novel approach that employs a window level KV cache selection method according to the specific requirements of the task, as shown in Figure~\ref{fig:model} (d). Unlike previous token level methods, our window selection method enhances semantic coherence in long-context inputs, by dynamically adapting to the task's characteristic, prioritizing relevant context, and closely mimicking human information processing. Additionally, to balance performance and efficiency, we integrate an intra-group layer KV cache indices sharing strategy.

\subsubsection{Task-Adaptive Window Selection}\label{sec:window_selection}
In our method, we retain the last $\alpha$ tokens as an observation window, while the remaining context, referred to as the review context, is divided into multiple review windows. The observation window is utilized to identify important review windows for caching across all layers, tailored to the specific task.

For clarity, we illustrate the attention mechanism using a single head. In a standard LLM, this is computed as follows:
\begin{equation}\label{eq1}
\mathbf{A} = \text{softmax} \left( \frac{\mathbf{Q} \cdot \mathbf{K}^{\mathsf{T}}}{\sqrt{d_k}} \right),
\end{equation}
where $d_k$ denotes the dimension of the key states.

To assess the importance of each token within the review context, we compute the attention score for each token based on the attention it receives from the observation window. This is formally expressed as:

\begin{equation}\label{eq2}
t_j = \sum_{i \in [n - \alpha, n]} \mathbf{A}_{ij}, \quad j \in [0, n - \alpha],
\end{equation}
where $t_j$ represents the score of the $j$-th token in the review context, $n$ denotes the input context length.

However, this token selection approach disrupts semantic coherence. To preserve semantic coherence and accommodate the variability in human reading patterns across different tasks, we further propose a task-adaptive window selection method. 
Specifically, in question-answering tasks, which can be viewed as \textbf{Information Localization}, humans must comprehend the semantic content of the entire review window to give accurate responses. In contrast, for tasks such as summarization and code completion, which can be viewed as \textbf{Information Aggregation}, humans only need to focus on the most critical and relevant context within individual review windows.

The review windows in the context can be represented as follows:
\begin{equation}\label{eq3}
\mathbb{W}_k = \{ t_j, \cdots, t_{j + \omega - 1} \}, \quad k \in \left[1, \left\lceil \frac{n - \alpha}{\omega} \right\rceil \right],
\end{equation}
where $\omega$ denotes the review window size, window $\mathbb{W}_k$ consists of tokens $t_j, \cdots, t_{j + \omega - 1}$, and $\left\lceil \cdot \right\rceil$ denotes the ceiling function.

To facilitate this process, a task-adaptive classifier is trained, with detailed training procedures described in Appendix~\ref{sec:appendix_detail} and~\ref{sec:appendix_classifier}. Consequently, the scoring function for evaluating the importance of review windows in the context for a specific task is defined as follows:
\begin{equation}\label{eq4}
s_k = \frac{1}{\min(p, \omega)} \cdot \text{sum}(\text{Top-}p(\mathbb{W}_k)),
\end{equation}
where $\text{Top-}p(\mathbb{W}_k) = \{ t_0', t_1', \cdots, t_{p-1}' \}$ represents the selection of the $p$ tokens with the highest scores from the $w$ tokens in the window. When $p = \omega$, it aligns with the information localization task. When $p < \omega$, the scenario aligns with the information aggregation task. The task type, identified by the task-adaptive classifier from the input context, is used to invoke the corresponding window selection method. Based on the allocated budget, a subset of high-scoring windows is retained from the review context to maintain the desired budget of KV cache. The detailed budget allocation strategy is described in Section~\ref{sec:budget}.

However, performing review window selection at every layer is computationally expensive. 
Ma et al.~\cite{ma2024compressing} demonstrate that the attention scores of adjacent layers in LLMs exhibit similarity. Additionally, Liu et al. ~\cite{liu2025chunkkv} proposed a layer-wise index reuse method under fixed layer budgets, which further validates the inter-layer similarity in LLMs.
Inspired by the inter-layer similarity characteristics of LLMs, we propose the intra-group layer KV cache indices sharing strategy to optimize the trade-off between performance and efficiency in WindowKV.

\subsubsection{Efficient Intra-Group Layer KV Cache Indices Sharing Strategy}\label{sec:group}
To enhance the efficiency of review window selection, an intra-group layer KV cache indices sharing strategy is employed.

Assume that the transformer layers of a model are denoted as \( \mathbb{L}= \{ l_0, l_1, \dots, l_{m-1} \} \), where \( m \) represents the number of layers in the model. The layers in \( \mathbb{L} \) are divided into $H = \frac{m}{\gamma }$ groups, each containing \( \gamma \) layers. For a given group \( \mathbb{G} = \{ l_{\text{g}}, l_{g+1}, \dots, l_{g + \gamma - 1} \} \), we apply the method from Section~\ref{sec:window_selection} to perform task-adaptive window selection on the first layer \( l_{\text{g}} \), obtaining the KV cache indices \( \mathbb{I}_{l_{\text{g}}} \) to be retained for that layer. For the remaining layers in the group \( \{ l_{g+1}, \dots, l_{g + \gamma - 1} \} \), they share the same KV cache indices \( \mathbb{I}_{l_{\text{g}}} \) as \( l_{\text{g}} \).

By adopting this approach, the computational cost can be significantly reduced, as the window selection algorithm is executed only once per group.

\subsubsection{Dynamic Budget Allocation}\label{sec:budget}
Inspired by PyramidKV~\cite{cai2024pyramidkv}, we allocate budgets to each group using an arithmetic sequence. The total budget for all groups is defined as:
\begin{equation}
\label{eq5}
{b^{\text{total}}} = \sum_{h \in [0, H - 1]} b^h,
\end{equation}
where \(H\) represents the number of groups.

For all groups \(\{ {\mathbb{G}_0}, \cdots ,{\mathbb{G}_{H - 1}}\} \), we first compute the budgets for the top group \({\mathbb{G}_{H - 1}}\) and the bottom group \({\mathbb{G}_0}\) as:

\begin{equation}
\label{eq6}
{b^{H - 1}} = \frac{{b^{\text{total}}}}{\lambda \cdot H} \quad \text{and} \quad {b^0} = \frac{2 \cdot {b^{\text{total}}}}{H} - {b^{H - 1}},
\end{equation}
where \(\lambda\) is a hyperparameter that controls the shape of the pyramid. The budgets for the remaining groups are calculated using the following equation:
\begin{equation}
\label{eq7}
{b^h} = {b^0} - \frac{{b^0 - {b^{H - 1}}}}{{H - 1}} \times h.
\end{equation}
Finally, the budget for each group is averagely distributed across all layers within the group.

\section{Experiments}\label{sec:exp}
\subsection{Experimental Setup}
\subsubsection{Backbone LLMs \& Benchmarks}
Due to computational constraints, we evaluate WindowKV against baseline methods using state-of-the-art open-source LLMs, specifically Qwen2.5-1.5B-Instruct~\cite{yang2024qwen2} and LLaMA3-8B-Instruct~\cite{touvron2023llama}. 
LongBench~\cite{bai-etal-2024-longbench} and Needle-in-a-Haystack~\cite{fu2024data} are two widely used benchmarks for evaluating KV cache compression methods. LongBench is specifically designed to assess the ability of language models to handle extended contexts. 
Needle-in-a-Haystack evaluates a model’s ability to locate key information within long input sequences, testing the in-context retrieval capabilities of LLMs across various KV cache compression methods.

\begin{table*}[h]
\centering

\resizebox{0.87\textwidth}{!}{
\begin{tabular}{l@{\hspace{0.05ex}}c@{\hspace{0.05ex}}c@{\hspace{0.05ex}}c@{\hspace{0.05ex}}c@{\hspace{0.05ex}}c@{\hspace{0.05ex}}c@{\hspace{0.05ex}}c@{\hspace{0.05ex}}c@{\hspace{0.05ex}}c@{\hspace{0.05ex}}c@{\hspace{0.05ex}}c@{\hspace{0.05ex}}c@{\hspace{0.05ex}}c@{\hspace{0.05ex}}c@{\hspace{0.6ex}}c@{\hspace{0.6ex}}c@{\hspace{0.6ex}}c}

\specialrule{1pt}{0pt}{2pt}
\multirow{6}{*}{\textbf{Method}} & \multicolumn{6}{c}{\textbf{Information Localization Task}} & \multicolumn{10}{c}{\textbf{Information Aggregation Task}} & \multirow{5}{*}{\textbf{Avg.}} \\
\cmidrule(lr){2-7} \cmidrule(lr){8-17}
& \multicolumn{3}{c}{Single-Document QA} & \multicolumn{3}{c}{Multi-Document QA}& \multicolumn{3}{c}{Summarization}& \multicolumn{3}{c}{Few-shot Learning}& \multicolumn{2}{c}{Synthetic} & \multicolumn{2}{c}{Code} &  \\
\cmidrule(lr){2-4}\cmidrule(lr){5-7}\cmidrule(lr){8-10}\cmidrule(lr){11-13}\cmidrule(lr){14-15}\cmidrule(lr){16-17}
& \rotatebox[origin=c]{30}{NrtvQA} & \rotatebox[origin=c]{30}{Qasper} & \rotatebox[origin=c]{30}{MF-en} & \rotatebox[origin=c]{30}{HotpotQA} & \rotatebox[origin=c]{30}{2WikiMQA} & \rotatebox[origin=c]{30}{Musique} & \rotatebox[origin=c]{30}{GovReport} & \rotatebox[origin=c]{30}{QMSum} & \rotatebox[origin=c]{30}{MultiNews} & \rotatebox[origin=c]{30}{TREC} & \rotatebox[origin=c]{30}{TriviaQA} & \rotatebox[origin=c]{30}{SAMSum} & \rotatebox[origin=c]{30}{PCount} & \rotatebox[origin=c]{30}{PRe} & \rotatebox[origin=c]{30}{Lcc} & \rotatebox[origin=c]{30}{RB-P} & \\ 

\cmidrule(lr){2-17}
& F1 & F1 & F1 & F1 & F1 & F1 & R-L & R-L & R-L & Acc (CLS) & F1 & R-L & Acc (EM) & Acc (EM) & Edit Sim & Edit Sim \\

\midrule
\multicolumn{18}{c}{Qwen2.5-1.5B-Instruct, Max Input Length = 15500, KV Size = Full} \\
\arrayrulecolor{black!20}\midrule
FKV &17.51 & 25.44 & 41.80 & 39.33 & 32.89 & 20.05 & 28.17 & 20.49 & 21.07 & 68.00 & 81.70 & 39.05 & 2.17 & 19.50 & 37.02 & 43.90 & 33.63\\

\arrayrulecolor{black!20}\midrule
\multicolumn{18}{c}{Qwen2.5-1.5B-Instruct, Max Input Length = 15500, KV Size = 2048} \\
\arrayrulecolor{black!20}\midrule
SLM & 14.41 & 20.61 & 29.97 & 31.27 & 31.28  & 13.38  & 20.44 & 18.81  & 20.63  & 63.00 & \textbf{81.00} & 37.72 & \textbf{3.50}                                 & 12.00            & \textbf{37.26} & 42.94 & 29.89 \\
H2O & 17.06   & 23.58 & 38.14  & 36.83  & 31.94   & 17.66  & \textbf{24.07} & 18.62   & \textbf{21.18} & 67.50  & 78.21   & 35.73  & 3.00                         & 11.06           & 36.27 & 40.66 & 31.34\\
PKV & 16.93  & 21.66 & 39.32  & \textbf{40.23} & \textbf{32.08} & 19.46  & 20.64   & 19.36  & 20.80  & 67.00   & 80.69  & \textbf{37.75} & 2.56                    & \textbf{18.50} & 36.11 & 41.71 & 32.18\\
Ours & \textbf{17.50} & \textbf{25.39} & \textbf{41.83} & 39.86  & 31.44 & \textbf{20.99} & 19.09 & \textbf{20.71} & 20.09  & \textbf{67.50} & 80.99 & 37.13& 3.00 & 17.00          & 36.98 & \textbf{44.53} & \textbf{32.75}\\

\arrayrulecolor{black!20}\midrule
\multicolumn{18}{c}{Qwen2.5-1.5B-Instruct, Max Input Length = 15500, KV Size = 1024} \\
\arrayrulecolor{black!20}\midrule
SLM & 12.01 & 13.50 & 24.69 & 29.30 & 29.68 & 11.07 & 17.17 & 16.80 & 19.20 & 56.50 & 78.34 & \textbf{37.79} & 3.00 & 8.00 & 36.82 & 40.95 & 27.18 \\
H2O & 16.20 & 21.02 & 33.60 & 35.96 & 30.69 & 15.62 & \textbf{22.42} & 18.68 & \textbf{20.37} & 67.00 & 75.45 & 32.98 & 3.00 & 11.05 & 34.87 & 38.67 & 29.85 \\
PKV & \textbf{16.79} & 19.41 & 38.01 & 39.25 & \textbf{32.23} & 17.12 & 18.06 & 18.66 & 19.14 & 66.00 & \textbf{80.32} & 36.78 & 2.62 & 14.08 & 36.67 & 39.11 & 30.89 \\
Ours & 16.63 & \textbf{22.12} & \textbf{42.16} & \textbf{39.49} & 31.56 & \textbf{19.63} & 15.00 & \textbf{19.61} & 18.71 & \textbf{67.50} & 79.40 & 36.77 & \textbf{3.00} & \textbf{14.50} & \textbf{37.48} & \textbf{41.79} & \textbf{31.58} \\

\arrayrulecolor{black!20}\midrule
\multicolumn{18}{c}{Qwen2.5-1.5B-Instruct, Max Input Length = 15500, KV Size = 512} \\
\arrayrulecolor{black!20}\midrule
SLM  &11.35 &11.89 &23.58 &27.79  &27.62 &11.05  &14.67  &16.75 &16.68  &50.50  &76.10 &\textbf{37.89}                                &3.00         &6.00          &35.60 &39.23 &25.61\\
H2O  &15.37 &\textbf{22.50} &30.45   &36.39  &29.42  &16.54  &\textbf{20.56} &17.82 &\textbf{19.60} &\textbf{66.00} &74.40  &31.45    &3.00         &6.92          &33.28 &34.81 &28.66\\
PKV  &\textbf{16.98} &17.84  &36.25  &37.87  &\textbf{30.61} &15.82  &16.17 &17.88  &16.89  &64.50 &78.80  &35.00                     &2.62         &\textbf{12.50} &34.72 &36.69 &29.45\\
Ours &16.32 &17.57 &\textbf{40.05} &\textbf{38.26} &30.03 &\textbf{17.18} &12.58 &\textbf{19.18} &15.79 &63.50 &\textbf{78.85} &36.96 &\textbf{3.00} &11.00         &\textbf{36.89} &\textbf{41.59} &\textbf{29.92}\\

\arrayrulecolor{black}\midrule
\multicolumn{18}{c}{LLaMA3-8B-Instruct, Max Input Length = 7950 , KV Size = Full} \\
\arrayrulecolor{black!20}\midrule

FKV& 25.59 & 32.04 & 39.67 & 43.61 & 35.29 & 21.30 & 28.64 & 23.15 & 26.69 & 71.50 & 90.48 & 42.59 & 4.86 & 69.75 & 56.84 & 52.16 & 41.51 \\

\arrayrulecolor{black!20}\midrule
\multicolumn{18}{c}{LLaMA3-8B-Instruct, Max Input Length = 7950 , KV Size = 2048} \\
\arrayrulecolor{black!20}\midrule
SLM  &24.00&24.00&30.18  &39.03  &31.59  &17.82 &24.92&21.59   &26.30  &68.00                                  &89.62           &41.65   &5.58         &69.67          &\textbf{58.78} &\textbf{56.13} &39.30\\
H2O  &26.07&28.95 &37.19  &42.62 &32.97 &19.77 &\textbf{27.40} &22.71 &\textbf{26.65} &71.00                   &\textbf{90.93}  &42.13   &\textbf{5.88} &70.00         &57.52          &55.42 &41.08\\
PKV  &25.41 &\textbf{29.70} &38.62&43.37  &\textbf{35.83} &\textbf{21.97} &26.94 &\textbf{23.09} &26.10 &70.50 &90.56   &\textbf{42.37}  &5.13          &69.75         &57.88          &53.87 &41.32\\
Ours &\textbf{26.61} &29.27 &\textbf{38.95} &\textbf{44.26} &34.76&21.17 &25.47&22.89  &25.86 &\textbf{71.50}  &90.48           &41.29   &5.43          &\textbf{70.00} &58.08         &55.63 &\textbf{41.35} \\

\arrayrulecolor{black!20}\midrule
\multicolumn{18}{c}{LLaMA3-8B-Instruct, Max Input Length = 7950 , KV Size = 1024} \\
\arrayrulecolor{black!20}\midrule
SLM &20.94 & 14.33  & 29.49 & 39.29 & 29.44  & 16.01 & 23.27 & 21.12 & 25.95 & 67.00 & 84.44 & \textbf{41.33} & \textbf{5.87} & 70.00 & 58.16 & 53.32 & 37.50\\
H2O &25.15 & 27.26  & 35.05  & 42.71  & 30.48 & 18.91  & \textbf{26.28} & 22.80 & \textbf{26.07} & \textbf{70.50} & \textbf{91.21} & 41.07 & 5.55 & 69.53  & 57.77 & 54.85 & 40.32\\
PKV &\textbf{26.02} & \textbf{27.31} & 37.10  & \textbf{43.85} & \textbf{33.86} & \textbf{21.18} & 24.71 & \textbf{23.21} & 25.26 & 70.00  & 90.56  & 41.28  & 5.58 & 69.75 & 57.33 & 53.29 & \textbf{40.64}\\
Ours &25.48 & 24.04 & \textbf{39.41} & 43.42 & 32.70  & 20.96  & 23.90  & 22.41  & 25.14 & 69.00 & 89.84 & 40.81 & 5.68 & \textbf{70.00} & \textbf{58.58} & \textbf{56.79} &40.51\\

\arrayrulecolor{black!20}\midrule
\multicolumn{18}{c}{LLaMA3-8B-Instruct, Max Input Length = 7950 , KV Size = 512} \\
\arrayrulecolor{black!20}\midrule
SLM  &20.70 &12.14 &22.08 &35.14  &27.06 &15.54   &21.01 &20.92  &23.84 &60.50  &83.49  &40.32  &5.79  &68.22 &58.59  &53.36 &35.54 \\
H2O  &23.29 &20.89  &31.38  &40.47 &30.30  &17.50 &\textbf{24.71} &21.80  &\textbf{25.78} &67.50 &\textbf{90.67} &39.67 &5.81   &68.49 &\textbf{59.26} &54.63 &38.88 \\
PKV  &\textbf{24.25} &\textbf{23.19} &36.17  &43.06 &32.13 &\textbf{20.41} &23.31 &\textbf{22.48} &24.23 &\textbf{70.00} &90.61 &\textbf{40.79} &\textbf{5.83} &70.00 &57.13  &53.70 &\textbf{39.83} \\
Ours &23.74 &21.30 &\textbf{37.27} &\textbf{44.03} &\textbf{32.28} &20.26 &21.42 &22.19  &23.28 &63.50   &88.18 &40.55 &5.42 &\textbf{70.00} &58.65 &\textbf{57.25} &39.33\\

\arrayrulecolor{black}\bottomrule
\end{tabular}
}
\caption{Performance comparison of WindowKV (Ours) with StreamingLLM (SLM), H2O, PyramidKV (PKV), and FullKV (FKV) on LongBench. WindowKV achieves the highest number of state-of-the-art results across various backbone LLMs and KV cache sizes. The best performance is highlighted in bold text.}
\label{table:longbench}
\end{table*}

\subsubsection{Baseline Methods}\label{sec:baseline}

We compare WindowKV with three state-of-the-art methods: \textbf{StreamingLLM (SLM)}~\cite{xiaoefficient}, \textbf{Heavy Hitter Oracle (H2O)}~\cite{zhang2023h2o} and \textbf{PyramidKV (PKV)}~\cite{cai2024pyramidkv}, as well as the use of full KV. Among these, SLM and H2O allocate a uniform KV cache size across all layers, while PKV assigns different KV cache sizes to different layers. Each method adopts a distinct KV cache compression strategy. 
For more detailed information, please refer to Appendix~\ref{sec:appendix_baseline}.

\subsection{Analysis on Intra-Group Layer KV Cache Indices Sharing Strategy}\label{sec:analysis}
\begin{figure}[htbp]
    \centering
    \includegraphics[width=0.45\textwidth]{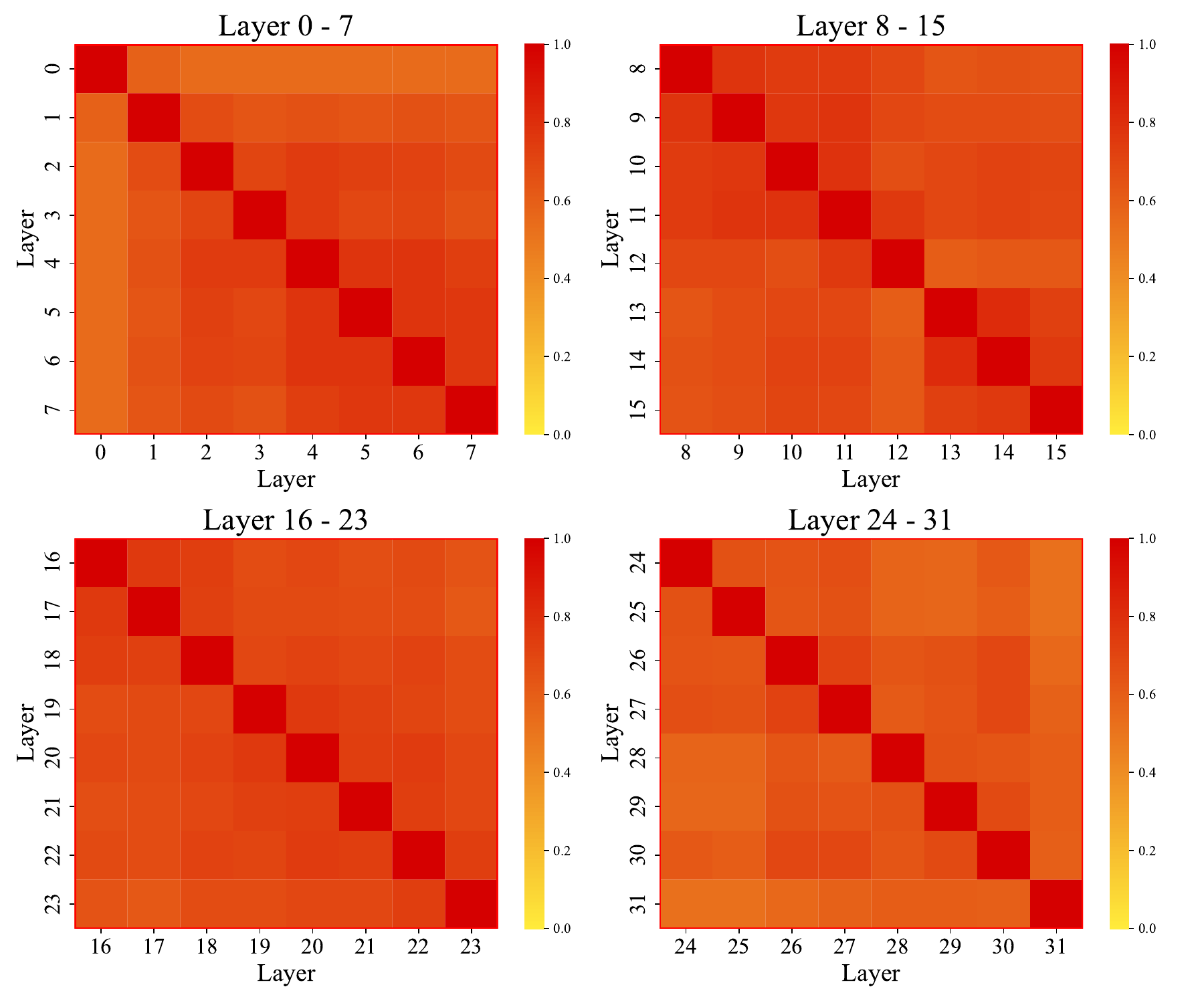}
    \caption{Similarity of Intra-Group Layer KV Cache Indices.}
    \label{fig:heatmap}
\end{figure}

In this section, we present some experiments to further analyze and validate the feasibility of intra-group layer KV cache indices sharing strategy.
First, we sampled multiple data points from each dataset in LongBench~\cite{bai-etal-2024-longbench} and conducted a comparative analysis on the similarity among intra-group layer KV cache indices. Using the budget allocation strategy outlined in Section~\ref{sec:budget}, we divided the 32 layers of LLaMA3-8B-Instruct into 4 groups, assigning distinct budgets to each group. After averagely distributing the budget across layers within each group, we applied the window selection method described in Section~\ref{sec:window_selection} to identify the retained windows and their corresponding KV cache indices for each layer. We then computed the Jaccard similarity of the KV cache indices between layers within the same group. The results, illustrated in Figure~\ref{fig:heatmap}, are presented in a heatmap where each cell represents the similarity between the retained KV cache indices of two layers within the same group. The experimental findings reveal that the KV cache indices of layers within the same group in WindowKV exhibit significant similarity, thereby validating the effectiveness of our KV cache indices sharing strategy.
For further analysis, please refer to Appendix~\ref{sec:appendix_analysis}.

\subsection{Main Results}\label{sec:main_result}
The evaluation results for LongBench are presented in Table~\ref{table:longbench}. We report the performance of Qwen2.5-1.5B-Instruct and LLaMA3-8B-Instruct across three KV cache sizes: 512, 1024, and 2048. As shown in Table~\ref{table:longbench}, the datasets in LongBench are categorized into two types: information localization task and information aggregation task. WindowKV achieves the highest number of state-of-the-art results across various backbone LLMs and KV cache configurations, demonstrating its superior adaptability and robustness across a wide range of tasks. 
The performance of WindowKV on some datasets is comparable to or slightly inferior to that of PyramidKV. This may be attributed to the fact that PyramidKV's token selection method, although disrupting semantic coherence, is able to consistently identify important tokens. 
In contrast, WindowKV maintains semantic coherence within a fixed-size window but may lose crucial tokens, making it challenging to achieve optimal performance across all datasets. 
Therefore, exploring adaptive review window size is a critical issue in our future work.
For implementation details, please refer to Appendix~\ref{sec:appendix_detail}.

\subsection{Results on Needle-in-a-Haystack}\label{sec:haystack}
\begin{figure}[htbp]
    \centering
    \includegraphics[width=0.48\textwidth]{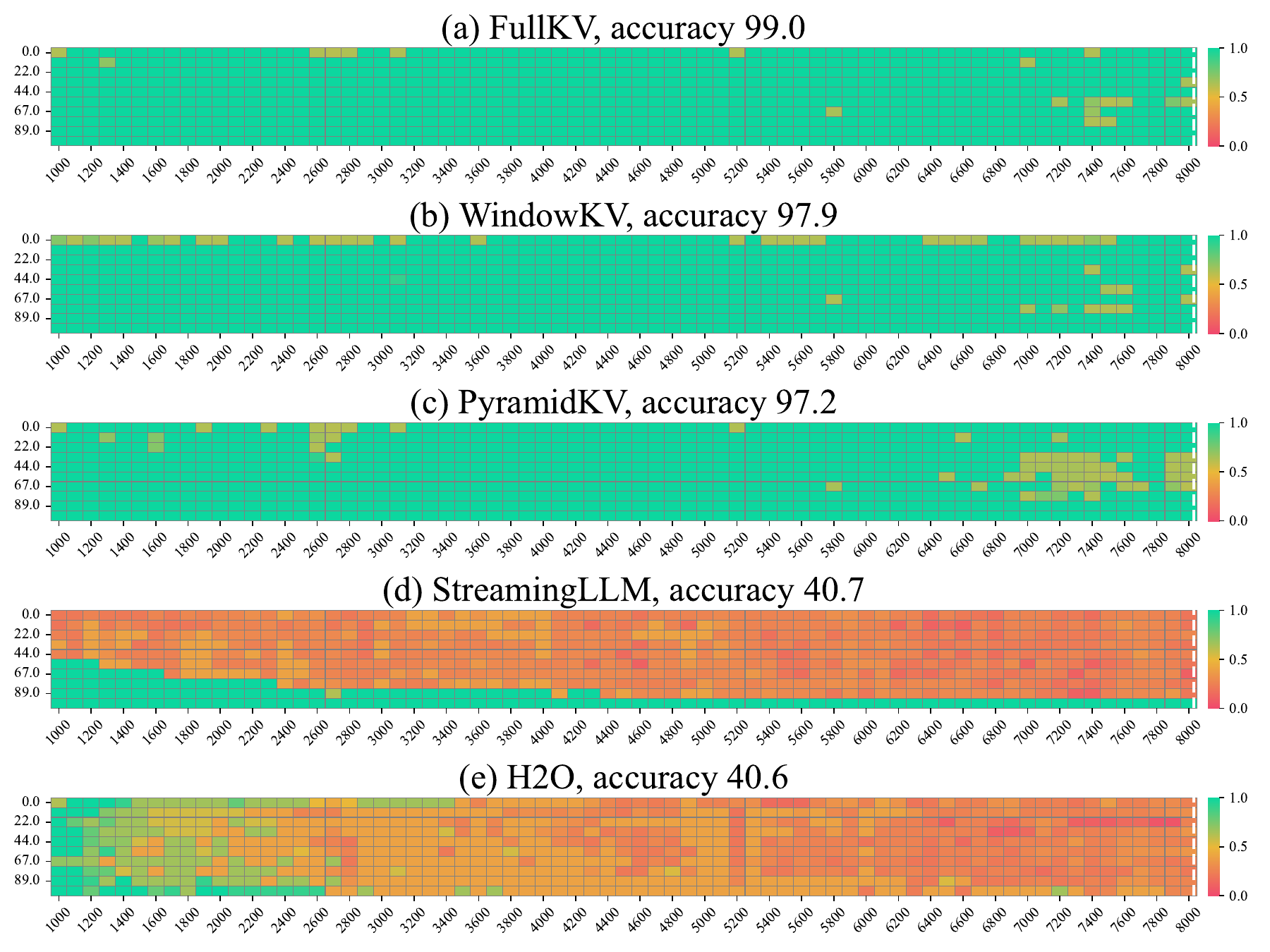}
    \caption{Needle-in-a-Haystack for LLaMA3-8B-Instruct with 512 KV cache size at 8K context length.}
    \label{fig:niah}
\end{figure}

We use Needle-in-a-Haystack to evaluate the long-context retrieval capabilities of LLMs. The Rouge-1 F1 metric is applied to assess the accuracy of the retrieved information. Several KV cache compression methods are evaluated. Figure~\ref{fig:niah} presents the benchmark results for LLaMA3-8B-Instruct, with the context length set to 8k tokens, which corresponds to the maximum length on the horizontal axis. The vertical axis represents the depth percentage. 
The results demonstrate that WindowKV outperforms other KV cache compression methods.

\subsection{Throughput Test}\label{sec:throughput}
Table~\ref{table:throughput} compares the throughput and latency of Vanilla, WindowKV, and WindowKV + Classifier. Compared to Vanilla, the Vanilla + WindowKV + Classifier configuration achieves a throughput increase of 117 tokens/s and a latency reduction of 0.17 ms/token. Moreover, the results indicate that incorporating the classifier does not significantly degrade efficiency.

\begin{table}[htbp]
\centering
\resizebox{\linewidth}{!}{
\begin{tabular}{l|c|c}
\toprule
Model &  Throughput (token/s) & Latency (ms/token) \\ \midrule
              Vanilla   & 764 &  1.31 \\ \midrule
              Vanilla  + WindowKV  & 894 &  1.12 \\ \midrule
              Vanilla  + WindowKV + Classifier & 881 & 1.14 \\
\bottomrule
\end{tabular}
}
\caption{Throughput test results. Vanilla refers to LLaMA3-8B-Instruct. The prefill length and generation length are 7,950 and 242, respectively. The experiment is conducted on a single A100 40G GPU with a KV cache size of 512 and repeated 10 times, with the results averaged.}
\label{table:throughput}
\end{table}

\subsection{Discussion and Analysis}\label{sec:discussion}
\begin{figure}[t]
    \centering
    \includegraphics[width=0.48\textwidth]{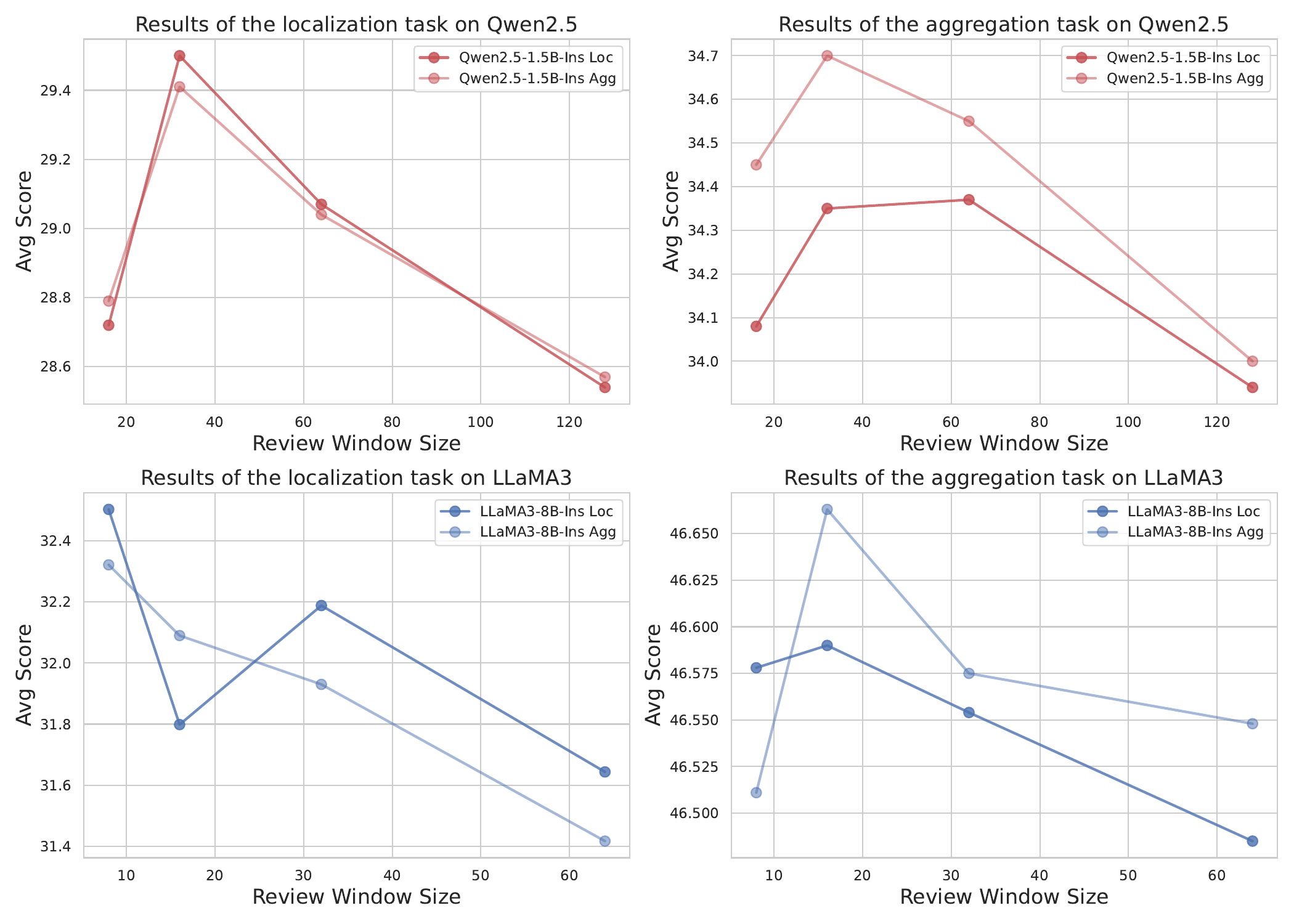}
    \caption{Impact of task-adaptive window selection and review window size on WindowKV performance.}
    \label{fig:zxt}
\end{figure}

This section examines the necessity of task-adaptive window selection method. When the task-adaptive classifier identifies the input context as an information localization task, an information localization-based window selection method outperforms an information aggregation-based approach, and vice versa, as shown in Figure~\ref{fig:zxt}. 
Additionally, the figure illustrates the impact of different review window sizes on WindowKV’s performance. In our experiments, the review window size varies among \{8, 16, 32, 64, 128\}. 
The Qwen2.5-1.5B-Instruct model achieves optimal performance across all tasks with a window size of 32. For the LLaMA3-8B-Instruct model, optimal performance is attained with a window size of 8 for information localization tasks and a window size of 16 for information aggregation tasks.

\section{Conclusion}
In this work, we present WindowKV, a method designed to address two issues in existing methods: preserving semantic coherence and considering task-specific characteristics during compression.
Evaluations on the LongBench demonstrate that WindowKV achieves performance comparable to full KV cache retention while using only 12\% of the original KV cache, significantly reducing memory requirements. Moreover, it outperforms other baselines in the Needle-in-a-Haystack test.

\section*{Ethical Considerations}
We highly prioritize ethical considerations and strictly adhere to the ACL Ethics Policy. In this paper, we propose a novel task-adaptive KV cache window selection method, WindowKV. WindowKV dynamically selects local semantic windows consisting of consecutive tokens, according to task-specific characteristics, ensuring the retained KV cache captures continuous, essential context. During the inference phase, WindowKV uses only 12\% of the original KV cache, significantly reducing memory requirements and improving inference speed. The methods and resources presented in this paper are open-source and widely used by researchers in the field of KV cache compression. The research results and conclusions presented in this paper are accurate and objective reports.

\bibliography{main}

\appendix

\section{Appendix}
\label{sec:appendix}

\subsection{Related Work}\label{sec:appendix_related_work}
\textbf{KV Cache Efficiency in Training and Inference}

\noindent Extensive previous research has explored methods for modifying transformer architecture to reduce KV cache size, including head-wise~\cite{shazeer2019fast, ainslie2023gqa, liu2024deepseek_v2} and layer-wise approaches~\cite{brandon2024reducing, sun2024you, wu-tu-2024-layer}. However, such modifications often require substantial computational resources for model retraining, making them less practical in settings with limited GPU resources.
An alternative line of research focuses on compressing the KV cache during the inference phase. Related approaches in this area include low-rank decomposition~\cite{dongget}, quantization~\cite{liu-etal-2024-intactkv}, token selection methods~\cite{liu2023scissorhands, zhang2023h2o, xiaoefficient, yang2024pyramidinfer, cai2024pyramidkv}. Among these methods, StreamingLLM~\cite{xiaoefficient} maintains a fixed KV cache size by retaining the KV states of both the initial and most recent tokens. Building on this idea, H2O~\cite{zhang2023h2o} and Scissorhands~\cite{liu2023scissorhands} use fixed-length KV cache size, selectively preserving the KV states of important tokens while evicting less critical ones.
Extending this line of research, PyramidInfer~\cite{yang2024pyramidinfer} and PyramidKV~\cite{cai2024pyramidkv} highlight that using the same KV cache size across all layers often leads to suboptimal performance. To address this issue, they propose a KV cache budget allocation strategy that assigns varying cache budgets to different layers, forming a pyramid structure.
Despite these advances, most existing methods rely on token-by-token selection of KV states, disrupting the semantic coherence of the context. This deviates from human reading behavior, where information is retrieved at the window level, rather than at the token level, particularly in long-context scenarios. In this work, we focus primarily on layer-wise window selection methods.

\noindent\textbf{Task-Adaptive Compression Methods}

\noindent In the field of model compression, TED~\cite{liang2023less} addresses the challenge of layer-wise distillation by introducing task-aware filters that align the hidden representations of the student and teacher models. These filters extract task-relevant knowledge, reducing the knowledge gap and enabling the student model to better adapt to the target task.
In the field of prompt compression, 
Style-Compress~\cite{pu2024style} argues that different tasks favor compressed prompts in distinct styles (e.g., extractive or abstractive), and optimizing compression performance requires identifying the most effective style for each task. Building on this insight, they introduce Style-Compress, a lightweight framework that enables smaller models to compress prompts for larger models across various downstream tasks without requiring additional training.
Moreover, TACO-RL~\cite{shandilya2024taco} critiques existing compression techniques for relying on suboptimal metrics or treating the task as task-agnostic. It proposes a novel task-aware method using reinforcement learning with task-specific rewards, guided by the lightweight REINFORCE algorithm.
However, task-adaptive approaches in KV cache compression field remain unexplored.

\subsection{Baselines}
\label{sec:appendix_baseline}
\noindent\textbf{SLM} maintains efficient long-context modeling by enabling LLMs trained with finite attention windows to generalize to infinite sequence lengths without fine-tuning. It leverages the attention sink phenomenon, where preserving the KV states of initial tokens largely restores the performance. 
In our experiments, for consistency with other methods, SLM retains the KV cache for the most recent \(\alpha\) tokens and the initial \(b - \alpha\) tokens, where \(b\) denotes the per-layer KV cache size.

\noindent\textbf{H2O} enhances KV cache efficiency by dynamically balancing the retention of recent tokens and Heavy Hitter (H2) tokens. It is based on the observation that a small subset of tokens contributes to most of the attention scores. 
H2O maintains the KV cache for the most recent tokens and the identified H2 tokens, where the eviction policy is guided by average attention scores computed across all queries.

\noindent\textbf{PKV} enhances KV cache management by dynamically adjusting the cache size across layers, leveraging the pyramidal information funneling effect in LLMs. 
PKV allocates more KV cache to lower layers and less to higher layers, deviating from previous approaches that use a uniform cache size. Furthermore, instead of aggregating attention across all queries, PKV captures attention signals based on patterns from instruction tokens, enabling more targeted and efficient compression.

\noindent\textbf{Full KV (FKV)} serves as the upper bound. It stores all keys and values for every token at every layer. All other methods need to be compared with Full KV.

\subsection{Implementation Details}\label{sec:appendix_detail}
In our method, for all tasks in LongBench, we use the prompts recommended by LongBench and follow its standard evaluation metrics~\cite{bai-etal-2024-longbench}. 
To eliminate variability introduced by sampling-based decoding, we employ greedy decoding for answer generation in both the Qwen2.5-1.5B-Instruct~\cite{yang2024qwen2} and the LLaMA3-8B-Instruct~\cite{touvron2023llama} model.
Specifically, for Qwen2.5-1.5B-Instruct~\cite{yang2024qwen2}, the number of shared layers is set to 7, and the review window size to 32. The observation window size for the information localization and aggregation tasks are 4 and 16, respectively. 
Due to computational constraints, the maximum input length for Qwen2.5-1.5B-Instruct is limited to 15,500 tokens. 
For LLaMA3-8B-Instruct~\cite{touvron2023llama}, the number of shared layers is set to 8.
For the information localization task, the review window size and observation window size are 8 and 16. For the information aggregation task, the review window size and observation window size are 16 and 32.
\(\lambda\) is used to control the shape of the pyramid, and it is 14 for all experiments.
Additionally, the classifier is trained on a dataset created by us, which consists of 9,551 samples divided into training, validation, and test sets with a ratio of 8:1:1.
The task-adaptive classifier is based on the bert-base-cased model and is trained with the following hyperparameters: batch size = 16, learning rate = 1e-6, dropout rate = 0.5, and 10 epochs. 
Experiments are conducted using 8× A100 GPUs with 40 GB of memory.

\subsection{Supplementary Analysis on Intra-Group Layer KV Cache Indices Sharing}\label{sec:appendix_analysis}
We conducted an experimental evaluation of the performance of WindowKV with various shared layer configurations on the Qwen2.5-1.5B-Instruct and LLaMA3-8B-Instruct models, as detailed in Table~\ref{table:shared_layer}. The results indicate that performance variations are negligible when the number of shared layers is set to 1, 4, or 7 for Qwen2.5-1.5B-Instruct. However, a significant decline in performance is observed when the number of shared layers is increased to 14 for Qwen2.5-1.5B-Instruct. 
Moreover, according to Equation~\eqref{eq5}-\eqref{eq7}, under a fixed total budget, as the number of shared layers increases, the budget allocated to each group becomes more evenly distributed. Specifically, when the number of shared layers is set to 1, the budget distribution is lopsided, resulting in more budget allocated to the earlier layers and significantly less to the later layers. Conversely, when the number of shared layers is set to 14 for Qwen2.5-1.5B-Instruct (and 16 for LLaMA3-8B-Instruct), the budget becomes overly even, disrupting the pyramid-shaped distribution. To balance performance and computational efficiency, we opted for 7 shared layers for Qwen2.5-1.5B-Instruct (and 8 for LLaMA3-8B-Instruct). This configuration ensures both the preservation of the pyramid-shaped distribution and a evenly budget allocation across layers.

\begin{table}[htbp]
  \centering
  \resizebox{1\linewidth}{!}{
  \begin{tabular}{c|c|c|c|c|c|c}
    \toprule
    \multirow{2}{*}{WindowKV, KV Size = 2048} & \multicolumn{6}{c}{LongBench Avg Score} \\ \cmidrule{2-7}
                        & \( \gamma=1 \) & \( \gamma=4 \) & \( \gamma=7 \) & \( \gamma=8 \) & \( \gamma=14 \) & \( \gamma=16 \)  \\ \midrule
    Qwen2.5-1.5B-Instruct  &32.13 &32.40 &\textbf{32.75} &- &27.83 &-\\ \midrule
    LLaMA3-8B-Instruct   &40.93 &40.78 &- &\textbf{41.35} &- &40.67\\ 
    \bottomrule
  \end{tabular}
      }
  \caption{LongBench performance with different layer-sharing scales, where $\gamma$ denotes the number of layers shared in each group.}
  \label{table:shared_layer}
\end{table}

\subsection{Effect of the Task-adaptive Classifier}\label{sec:appendix_classifier}
The task-adaptive classifier analyzes the input context to determine whether it corresponds to an information localization or aggregation task. 
The evaluation results for LongBench~\cite{bai-etal-2024-longbench}, as illustrated in Table~\ref{table:classifier}, indicate that the task-adaptive classifier achieves high accuracy with simple fine-tuning.

\begin{table}[htbp]
\centering
\resizebox{0.7\linewidth}{!}{
\begin{tabular}{c|c|c|c}
\toprule
\multirow{2}{*}{Model} & \multicolumn{3}{c}{LongBench} \\ \cmidrule{2-4} 
                       & Acc & Recall & F1 \\ \midrule
              bert-base-cased  & 92.69 & 95.19 &  94.75 \\
\bottomrule
\end{tabular}}
\caption{Classifier Test Result.}
\label{table:classifier}
\end{table}

\subsection{Limitations}
This study is limited to investigating layer-wise KV cache compression and does not explore head-wise approaches, which represent another highly active and promising research direction. Future work should extend this research by investigating head-wise compression techniques. Additionally, while the current study focuses on long-context input scenarios with compression applied exclusively during the prefilling phase, subsequent research could expand the scope to include KV cache compression in long-output generation scenarios.

\appendix
\end{document}